\documentclass{article}
\usepackage[preprint]{colm2026_conference}

\usepackage{microtype}
\usepackage{hyperref}
\usepackage{url}
\usepackage{booktabs}
\usepackage{lineno}
\usepackage{amsmath}
\usepackage{amssymb}
\usepackage{graphicx}
\usepackage{pgfplots}
\pgfplotsset{compat=1.17}
\usepackage{tikz}
\usetikzlibrary{shapes.geometric,arrows.meta,positioning,calc}
\usepackage{multirow}
\usepackage{xcolor}
\usepackage{array}
\usetikzlibrary{patterns}
\usepgfplotslibrary{groupplots}

\definecolor{darkblue}{rgb}{0,0,0.5}
\hypersetup{colorlinks=true,citecolor=darkblue,linkcolor=darkblue,urlcolor=darkblue}

\title{PermDoRA - Understanding Adapter Interference in \\ Language Models: Limits of Parameter-Space Geometry}

\author{Gowtham Sivaramakrishnan \\
Independent Researcher \\
\texttt{gowtham\_s@berkeley.edu}
\And
Sarvesha Kumar Kombaiah Seetha \\
Independent Researcher \\
\texttt{skombaiah@my.harrisburgu.edu}
\AND
Kishan Gupta Balaji \\
Independent Researcher \\
\texttt{kbalaji@my.harrisburgu.edu}
\And
Santhosh Baradwaj Vaduvur Ranganathan \\
Independent Researcher \\
\texttt{vaduvurr@usc.edu}
}

\begin{document}

\ifcolmsubmission
\linenumbers
\fi

\maketitle

\begin{abstract}
Access control in large language models (LLMs) requires modular mechanisms to enable domain-specific behavior without retraining or cross-domain interference. A common hypothesis is that interference during adapter composition arises from overlap in linear parameter updates, suggesting that enforcing orthogonality or directional independence should improve multi-domain performance.
We test this hypothesis using DoRA-RBAC, a hierarchical adapter composition framework based on weight-decomposed low-rank adaptation. We compare conventional Euclidean merging with a geometry-aware Riemannian-inspired merging strategy that approximates the Fréchet mean via normalized directional averaging across multiple QA benchmarks (GPQA, PubMedQA, SimpleQA, WMDP) on LLaMA-3.1-8B and Mistral-7B.
Our results show that while single-domain performance matches LoRA, geometry-aware merging provides no consistent advantage over standard averaging in multi-domain settings. Diagnostic analysis further reveals that angular alignment and orthogonality of adapter updates are weak predictors of composition performance.
These findings suggest that adapter interference is not governed primarily by parameter-space geometry, but is instead consistent with interactions in shared nonlinear representations.
\end{abstract}

\section{Introduction}
\label{sec:intro}
Large language models are increasingly deployed in settings where users have unequal access to domain-specific knowledge. Enterprise and scientific systems often combine multiple domains, while individual users are authorized to access only subsets of this information. This creates a key challenge: enabling modular domain adaptation while supporting controlled composition of multiple domains at inference time.

Existing access-control approaches operate at the prompt, routing, or retrieval level. While effective, they do not enforce modularity at the level of learned representations. This motivates a parameter-level perspective, where domain knowledge is stored in separate adapters that can be selectively activated and composed based on user permissions.

A central question in this setting is what drives interference when multiple adapters are composed. A common hypothesis is that interference arises from overlap in linear parameter updates, suggesting that enforcing orthogonality or directional separation should improve composition. Under this view, geometry-aware merging methods should outperform standard Euclidean averaging.

We test this hypothesis using DoRA-RBAC, a hierarchical adapter framework built on weight-decomposed~\citep{Liu+2024} low-rank adaptation (DoRA), which separates updates into direction and magnitude. Each domain is represented by a manager adapter on top of a shared baseline, and authorized subsets of domains are composed into a single flattened adapter at inference time.

This setup enables a controlled comparison between two composition strategies: (1) Euclidean merging via averaging and SVD refactorization, and (2) a geometry-aware, Riemannian-inspired merge that approximates the Fréchet mean~\citep{Karcher1977,Shoemake1985} through normalized directional averaging with separate magnitude aggregation. If interference is primarily driven by linear overlap, the geometry-aware approach should provide a consistent advantage.

Our results do not support this hypothesis. Across multiple QA benchmarks on Llama-3.1-8B~\citep{Grattafiori+2024} and Mistral-7B~\citep{Jiang+2023}, geometry-aware merging provides no consistent improvement over Euclidean merging in multi-domain settings. While angular alignment explains when geometry-aware merging collapses to Euclidean behavior, it does not predict overall composition performance. These findings indicate that enforcing orthogonality in parameter space is insufficient to resolve adapter interference and are instead consistent with interaction effects in shared nonlinear representations.

Finally, we evaluate implications for access control. Domain-specific adapters improve target-domain performance (e.g., +0.1178 UGI on SimpleQA ~\citep{Wei+2024simpleqa}, but distinguishability remains above chance DDI AUC $\approx$ 0.65, indicating that modular composition does not provide strong privacy guarantees. DoRA-RBAC should therefore be viewed as a modular composition framework rather than a mechanism for strict isolation.

\section{Related work}
\label{sec:related}

Adapters, prefix tuning ~\citep{Li+Liang2021}, and LoRA ~\citep{Hu+2022} enable task-specific fine-tuning by updating a small fraction of parameters. DoRA ~\citep{Liu+2024} extends LoRA by decomposing updates into direction and magnitude, enabling finer-grained analysis of adapter interactions during composition.

Model soups~\citep{Wortsman+2022}  and TIES-Merging ~\citep{Yadav+2023}combine fine-tuned models in a shared parameter space. Our setting is more constrained: we compose domain-specific low-rank updates selected by user access rights into a single adapter at inference time. Unlike prior merging work, composition is governed by an explicit access-control policy rather than task similarity.

PermLLM ~\citep{Jayaraman+2025} formalizes access-control enforcement over LLM responses; AdapterSwap ~\citep{Fleshman+2025} removes adapters based on access policies. Differentially private fine-tuning ~\citep{Yu+2022dp} bounds training-data leakage but does not address modular composition. Our work bridges these areas: we modularize domain adaptations and assess distinguishability via membership-inference metrics rather than providing formal privacy guarantees.

\section{Method}
\label{sec:method}

\subsection{Problem Setting and Hierarchical DoRA-RBAC}
We freeze a pretrained backbone and train a shared baseline adapter. For each domain $d$, a manager adapter stores an incremental update to this baseline. Following DoRA LoRA~\citep{Hu+2022},we decompose each manager's update into a unit-direction vector and scalar magnitude (Eq.~1), rather than the standard LoRA parameterization ${\Delta W}$ = $BA$. At inference time, only the manager adapters corresponding to a user's authorized domains are composed into a single flattened adapter.

In the final component of our framework DoRA-RBAC, each domain-specific manager adapter stores a unique update to the common baseline adapter, while the common adapter serves as the repository for all shared domains. When a user accesses an authorized domain for inference, only the manager update of this user's domains will be used to generate an output from that manager. This composite output will be constructed using a single adapter for all manager outputs.
\begin{equation}
  \hat{V}_i = \frac{\Delta W_i}{\|\Delta W_i\|_F}, \qquad
  \Delta W_i = B_i A_i, \qquad
  \Delta W_i^{\mathrm{DoRA}} = m_i \cdot \hat{V}_i,
  \quad m_i = \|\Delta W_i\|_F
  \label{eq:dora_decomp}
\end{equation}
where $\hat{V}i \in \mathbb{R}^{d{\mathrm{out}} \times d_{\mathrm{in}}}$ is the
normalized direction of the low-rank update and $m_i \in \mathbb{R}_{>0}$ is the
scalar magnitude for domain $i$

Also, each manager adapter $\theta_i$ is trained by minimizing:
\begin{equation}
  \mathcal{L}(\theta_i)
  =
  \mathcal{L}_{\mathrm{task}}(\theta_i;\mathcal{D}_i)
  +
  \frac{\lambda}{2}
  \sum_j
  F_j
  \bigl(
    \theta_i^{(j)} - \theta_{\mathrm{base}}^{(j)}
  \bigr)^2
  \label{eq:ewc}
\end{equation}
where $F_j$ is the diagonal Fisher information estimated at
$\theta_{\mathrm{base}}$, and $\lambda > 0$ is the EWC regularization coefficient.
This design makes domain knowledge modular by construction: authorized domains are composed explicitly through manager adapters rather than implicitly through prompts, and inference cost does not scale with the total number of authorized domains.

\subsection{Authorized Adapter Composition}
\label{sec:merge}

A user is permitted to access a specified collection of domains. Denote the domains the user has access to as  $D \in \{D1,\ldots,Dn\}$. We will describe two methods of building the subordinate manager adapters for the user across the domains included in D.
\begin{enumerate}
\item \textbf {Euclidean Composition} We recompute each manager effective weight update from the shared manager and domain-specific manager and subsequently calculate the average of each effective weight in an Euclidean manner; we then use truncated singular value decomposition to re-refactor the resulting effective weight into a low number of dimensions. This approach is a standard method of deriving an acceptable implementation method.
\begin{equation}
  \Delta W_D^{\text{Euc}} = \operatorname{SVD}_r\!\Bigl(
    \tfrac{1}{|D|}\sum_{i\in D}\Delta W_i
  \Bigr)
  \label{eq:euclidean_merge}
\end{equation}

\item \textbf{Riemannian Composition} Since DoRA differentiates normalized direction from magnitude, we can use a geometric method to compose effective updates. We take the normalized directional effective updates as points on a unit hypersphere in $\mathbb{R}^3$, compute a normalized directional averaging of these directional effective updates, and compute the average of their magnitudes using a Euclidean metric; we then re-factor the resulting effective weight using SVD to generate the final merged effective weight. This construction arises from the idea that normalized direction is a geometrical object; therefore, calculating the mean of the normals directly on the unit hypersphere containing the normals conserves their geometrical structure better.
\begin{equation}
  \hat{v}_D^{\text{Riem}} = \frac{\sum_{i\in D}\hat{v}_i}{\bigl\|\sum_{i\in D}\hat{v}_i\bigr\|},
  \qquad
  m_D^{\text{Riem}} = \sum_{i\in D} m_i
  \label{eq:riemannian_merge}
\end{equation}
\end{enumerate}

\subsection{Evaluation Metrics}
\label{sec:ewc}

For single domain evaluations, we show both target loss and utility of domains when performing single-domain evaluations. Specifically, for SimpleQA, we provide an evaluation of the gains attributable to the domain-specific manager implemented by summation of all domain-specific gains into a single measure called the Utility Gap Index (UGI)\citep{Jayaraman+2025}. 
Let $f_D^M$ denote the model composed with domain manager $D$, and let
$\mathcal{D}_S$ be the evaluation dataset for domain $S$.
Define the accuracy of the model on domain $S$ as:
\begin{equation}
  \mathrm{Acc}(f, \mathcal{D}_S)
  =
  \mathbb{E}_{(q, y) \sim \mathcal{D}_S}
  \bigl[
    \mathbf{1}\{f(q) = y\}
  \bigr]
  \label{eq:accuracy}
\end{equation}

The Utility Gap Index (UGI) for domain $S$ is then defined as:
\begin{equation}
  \mathrm{UGI}(S)
  =
  \mathrm{Acc}(f_S^M, \mathcal{D}_S)
  -
  \mathrm{Acc}(f_{\mathrm{base}}, \mathcal{D}_S)
  \label{eq:ugi}
\end{equation}
 where $f_S^M$ denotes the model with the domain-specific manager for $S$ activated,
$f_{\mathrm{base}}$ is the baseline model without the domain manager, and
$\mathrm{UGI}(S) \in [-1, 1]$ measures the change in task accuracy attributable to
the domain-specific adapter. Positive values indicate improved domain utility,
while negative values indicate degradation.

With respect to evaluating security we also utilize the Domain Distinguishability Index (DDI)\citep{Jayaraman+2025}. The DDI measures how well a model trained using a certain domain can or cannot be distinguished between models that did not train on that domain. To quantify model distinguishability we compute an inference-based score for each of the active domains e.g. using the Min-K++ confidence score from our earlier findings in Zhang et al., 2025 and subsequently compute an area (AUC) under the receiver operating curve for the task of classifying “in-domain” versus “out-of-domain” instances. A DDI AUC equal to 0.5 denotes chance versus 1.0 denotes an identifiable fingerprint; meaning that if a DDI is near 0.5 then there is no identifying characteristics of the behavioral fingerprints of either set of models. The tables presented in this document will be based upon using Min-K++-based DDI AUC values; Log-based probabilitiy, loss and Zlib versions will all be found in the appendix. Additionally we will conduct analyses based upon cosine similarity and symmetric orthogonality losses between the Experiment 6 generated data as a basis to determine if the dataset separation is predictive of the quality of the dataset created through domain composition.
\begin{equation}
DDI(m) = \mathbb{E}_{S_i \subseteq \mathbb{S}, S_j \subseteq \mathbb{S}}[m(O^{(S_i,S_j)})]
\end{equation}
\begin{itemize}
    \item $DDI(m)$: The Domain Distinguishability Index defined for a given membership inference metric $m$. 
    \item $\mathbb{E}$: The expectation taken over all domain sets. 
    \item $\mathbb{S}$: The domain universe consisting of $n$ total security domains. 
    \item $S_i, S_j \subseteq \mathbb{S}$: Ordered pairs of domain sets from the universe $\mathbb{S}$ that have no overlap (i.e., $S_i \cap S_j = \emptyset$). 
    \item $m(\cdot)$: A given membership inference metric. 
    \item $O^{(S_i,S_j)}$: The output of a membership inference oracle $O$, specifically defined as $O(f_D^\mathcal{M}(q)|S_i, f_D^\mathcal{M}(q)|S_j)$ for all queries $q \sim \mathcal{D}_{S_i}$. 
\end{itemize}
For the trade-off analysis we will additionally compute KL divergence statistics for the merged adapter versus using a union-based average of the scores from those active domains to represent a true utility for each merged adapter created versus a single merged adapter. The utility represented by the union-based average is higher than for each individual domain, however, the run time will increase linearly with the number of active domains.
\subsection{Hypothesis: Geometry Predicts Equivalence Regimes, Not Performance}
\label{sec:hypothesis}

We study a narrower and falsifiable hypothesis than the claim that parameter-space
geometry determines composition quality. Specifically, we test whether angular alignment
between adapter updates predicts when Euclidean and geometry-aware merging should behave
similarly. Let $\Delta W_i$ and $\Delta W_j$ denote the effective updates for domains $i$
and $j$, respectively. Their cosine similarity is
\begin{equation}
\cos(\Delta W_i,\Delta W_j)
=
\frac{\langle \Delta W_i,\Delta W_j\rangle_F}
{\|\Delta W_i\|_F\,\|\Delta W_j\|_F}.
\end{equation}

Our hypothesis is that as adapter updates become more collinear, the difference between
Euclidean and Riemannian-inspired merging should shrink:
\begin{equation}
\cos(\Delta W_i,\Delta W_j)\uparrow
\quad \Longrightarrow \quad
\left|
\mathrm{Acc}{\mathrm{Riem}}(D)-\mathrm{Acc}{\mathrm{Euc}}(D)
\right|\downarrow .
\end{equation}

This is an equivalence-regime hypothesis rather than a performance hypothesis.
That is, angular alignment may explain when geometry-aware merging collapses to
Euclidean behavior, but it does not by itself imply improved downstream utility or reduced
interference. We test this prediction empirically in Section~5.
.
\section{Experimental Setup}
\label{sec:experiments}

\subsection{Datasets and Domains}

We evaluate on four benchmarks spanning science, biomedicine, open-domain factual QA, and safety-sensitive knowledge. GPQA \citep{Rein+2024} is partitioned into biology, chemistry, and physics. PubMedQA \citep{Jin+2019} is clustered into ten semantic groups. SimpleQA\citep{Wei+2024simpleqa} is divided into ten categories including art, geography, history, and science and technology. WMDP\citep{Li+2024wmdp} is partitioned into bio, chem, and cyber domains.
We use WMDP only for the Llama experiments in the main paper. For Mistral, we focus the cross-backbone comparison on GPQA, PubMedQA, and SimpleQA.

\subsection{Backbones and Training Protocol}
\label{sec:backbone}
The primary foundation for our study was Llama version 3.1-8B\citep{Grattafiori+2024}. Training of LoRA and DoRA adapters was accomplished using rank = 16 and scale = 8; however, a shared baseline adapter was trained and subsequently frozen. Then, per domain, manager adapters were trained while maintaining the backbone and baseline adapter fixed.

Also, we utilized Mistral version 7-B\citep{Jiang+2023} as our Cross-Backbone Validation model. Results for Mistral were provided in the main paper for the single domain and multidomain merging investigations, however, all deeper ablation, compute-tradeoff, security and diagnostic analysis was performed centered on Llama producing a cleaner and coherent comparison than would otherwise have been offered through comparisons made using both models.

\subsection{Experiments}
\label{sec:exp2}

Our evaluation has six components: (1) single-domain utility of DoRA-RBAC vs.\ LoRA, (2) multi-domain composition on 2- and 3-domain SimpleQA combinations, (3) cosine-similarity and orthogonality diagnostics from independent artifacts, (4) compute--utility trade-offs via KL divergence to a union reference, (5) domain distinguishability via DDI on Llama SimpleQA, and (6) ablation under strong EWC regularization.

\section{ Results}
\label{sec: Results}
\subsection{Single-Domain Utility}
Table 1 outlines single domain target loss (loss) and UGI values across the various Llama experiments conducted by either DoRA or LoRA on all assessed datasets (GPQA, PubMedQA, SimpleQA, and WMDP). The average differences between both methods are comparably low (loss difference $<0.01$) across the four respective domains. However, the analysis of results from the various experimental methodologies leads the researchers to determine that there is no overall performance difference between DoRA and LoRA when running the Mistral method. Regardless of whether running GPQA, the average difference in loss values was significantly greater when comparing DoRA (3.65) versus LoRA (3.33).

Thus, the essence of Table 1 is not that one method clearly dominates over the other nor does it imply that DoRA-rbac is any less effective than LoRA in the single domain context (which would call into question their contributions). The original objective was to illustrate that modular decomposition methods developed in the previous chapter would still provide sufficient utility within the single domain; therefore, it creates the opportunity to evaluate the actual impact of decreasing inference caused when compare to coincidental use in the same domain 

\begin{table}[ht]
\centering
\caption{Single‑domain performance (mean $\pm$ standard deviation across runs). Lower loss and higher UGI are better. WMDP is reported only for Llama in the main paper.}
\resizebox{\textwidth}{!}{%
\begin{tabular}{llcccccc}
\toprule
\textbf{Backbone} & \textbf{Dataset} & \textbf{\begin{tabular}[c]{@{}c@{}}LoRA\\ loss\\$\pm$\,SD\end{tabular}} & \textbf{\begin{tabular}[c]{@{}c@{}}DoRA\\ loss\\$\pm$\,SD\end{tabular}} & \textbf{\begin{tabular}[c]{@{}c@{}}$\Delta$\\ loss\\$\pm$\,SD\end{tabular}} & \textbf{\begin{tabular}[c]{@{}c@{}}LoRA\\ UGI\\$\pm$\,SD\end{tabular}} & \textbf{\begin{tabular}[c]{@{}c@{}}DoRA\\ UGI\\$\pm$\,SD\end{tabular}} & \textbf{\begin{tabular}[c]{@{}c@{}}$\Delta$\\ UGI\\$\pm$\,SD\end{tabular}} \\
\midrule
Llama   & GPQA     & $1.73\pm0.10$ & $1.73\pm0.09$ & $+0.00\pm0.01$ & $0.101\pm0.029$ & $0.085\pm0.048$ & $-0.016\pm0.013$ \\
Llama   & PubMedQA & $1.80\pm0.06$ & $1.80\pm0.07$ & $+0.00\pm0.01$ & $0.0046\pm0.0012$ & $0.0044\pm0.0012$ & $-0.0002\pm0.0008$ \\
Llama   & SimpleQA & $3.10\pm0.16$ & $3.11\pm0.17$ & $+0.00\pm0.02$ & $0.0114\pm0.0033$ & $0.0114\pm0.0036$ & $+0.0001\pm0.0008$ \\
Llama   & WMDP     & $1.47\pm0.04$ & $1.47\pm0.04$ & $+0.00\pm0.00$ & $0.112\pm0.065$ & $0.116\pm0.068$ & $+0.005\pm0.010$ \\
Mistral & GPQA     & $3.33\pm2.11$ & $3.65\pm2.58$ & $+0.32\pm0.45$ & $0.122\pm0.033$ & $0.126\pm0.060$ & $+0.005\pm0.068$ \\
Mistral & PubMedQA & $1.79\pm0.07$ & $1.79\pm0.06$ & $-0.00\pm0.01$ & $0.0058\pm0.0025$ & $0.0059\pm0.0020$ & $+0.0002\pm0.0007$ \\
Mistral & SimpleQA & $2.82\pm0.16$ & $2.83\pm0.16$ & $+0.01\pm0.03$ & $0.0149\pm0.0042$ & $0.0149\pm0.0049$ & $+0.0000\pm0.0020$ \\
\bottomrule
\end{tabular}%
}
\end{table}

\subsection{Multi-Domain Composition}
In Table 2, we sum all SimpleQA utility and AUC (Area Under Curve) for DDIs based upon Min-K++ for combinations between two domains and three domains of SimpleQA. For both backbones, Euclidean and Riemannian-based composition (composition utilizing Euclidean and Riemaniann geometries) is almost equivalent. For the case with Llama, the average SimpleQA score for two-domain combinations is approximately 0.0053 under the Merge Rules; three-domain combinations have an approximate SimpleQA score of 0.0053 under each Merge Rule. For the case with Mistral, the average SimpleQA scores for two-domain combinations is approximately 0.0123, and for three-domain combinations is also approximately 0.0123 with both Merge Rules. The differences in SimpleQA scores and their inconsistencies are so minor that we cannot reasonably conclude that Geometry-Aware Composition (to compose observably from the same geometric properties) produces a substantial or measurable impact in reducing interference due to linear directional overlap.

This is the primary empirical finding of this paper. Had linear directionality overlap been the primary explanation for interference, then a merge rule that honors DoRA's normalized directions would have produced a stronger empirical finding. The gain realized from merging is less related to having a more correctly structured geometric composition, and more related to the properties of modularity of composition.
\begin{table}[ht]
\centering
\caption{Multi-domain composition on SimpleQA.  Values are mean $\pm$ standard deviation across all domain combinations.  Accuracy is the target-domain SimpleQA score; DDI AUC uses the Min–K\%++ variant.}
\begin{tabular}{lllll}
\toprule
\textbf{Backbone} & \textbf{Domains} & \textbf{Merge} & \textbf{SimpleQA} & \textbf{DDI AUC} \\
\midrule
Llama   & 2 & Euclidean  & $0.00532\pm0.00114$ & $0.6105\pm0.0278$ \\
Llama   & 2 & Riemannian & $0.00532\pm0.00113$ & $0.6116\pm0.0290$ \\
Llama   & 3 & Euclidean  & $0.00527\pm0.00074$ & $0.6142\pm0.0258$ \\
Llama   & 3 & Riemannian & $0.00525\pm0.00076$ & $0.6147\pm0.0250$ \\
Mistral & 2 & Euclidean  & $0.01236\pm0.00250$ & $0.6102\pm0.0214$ \\
Mistral & 2 & Riemannian & $0.01227\pm0.00248$ & $0.6116\pm0.0208$ \\
Mistral & 3 & Euclidean  & $0.01234\pm0.00174$ & $0.6084\pm0.0307$ \\
Mistral & 3 & Riemannian & $0.01222\pm0.00174$ & $0.6054\pm0.0302$ \\
\bottomrule
\end{tabular}
\end{table}
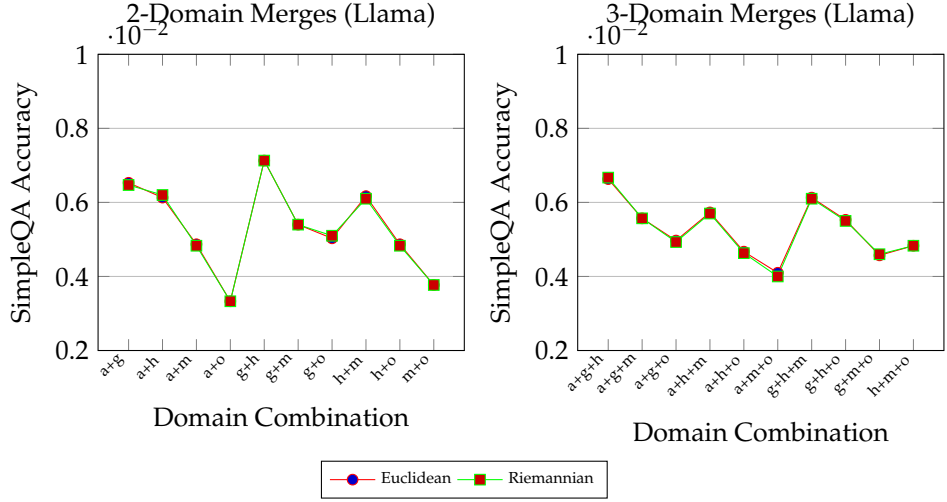
\begin{figure}[h]
\centering
\begin{tikzpicture}
\begin{groupplot}[
  group style={group size=2 by 1, horizontal sep=1.5cm},
  width=0.46\linewidth, height=5.5cm,
]
\nextgroupplot[
  title={2-Domain Merges (Llama)},
  xlabel={Domain Combination}, ylabel={SimpleQA Accuracy},
  xtick=data,
  xticklabels={
    a+g, a+h, a+m, a+o, g+h, g+m, g+o, h+m, h+o, m+o
  },
  xticklabel style={rotate=45,anchor=east,font=\tiny},
  legend style={at={(1,-0.5)},anchor=south,legend columns=2,font=\tiny},
  ymin=0.002, ymax=0.010,
  ymajorgrids=true,
  symbolic x coords={a+g,a+h,a+m,a+o,g+h,g+m,g+o,h+m,h+o,m+o},
]
\addplot+[mark=*,color=red] coordinates {
  (a+g,0.00653)(a+h,0.00613)(a+m,0.00487)(a+o,0.00333)
  (g+h,0.00713)(g+m,0.00540)(g+o,0.00503)(h+m,0.00617)(h+o,0.00487)(m+o,0.00377)
};
\addplot+[mark=square*,color=green] coordinates {
  (a+g,0.00647)(a+h,0.00620)(a+m,0.00483)(a+o,0.00333)
  (g+h,0.00713)(g+m,0.00540)(g+o,0.00510)(h+m,0.00610)(h+o,0.00483)(m+o,0.00377)
};
\legend{Euclidean, Riemannian}

\nextgroupplot[
  title={3-Domain Merges (Llama)},
  xlabel={Domain Combination}, ylabel={SimpleQA Accuracy},
  xtick=data,
  xticklabels={
    a+g+h, a+g+m, a+g+o, a+h+m, a+h+o, a+m+o,
    g+h+m, g+h+o, g+m+o, h+m+o
  },
  xticklabel style={rotate=45,anchor=east,font=\tiny},
  ymin=0.002, ymax=0.010,
  ymajorgrids=true,
  symbolic x coords={a+g+h,a+g+m,a+g+o,a+h+m,a+h+o,a+m+o,g+h+m,g+h+o,g+m+o,h+m+o},
]
\addplot+[mark=*,color=red] coordinates {
  (a+g+h,0.00663)(a+g+m,0.00557)(a+g+o,0.00497)(a+h+m,0.00573)
  (a+h+o,0.00467)(a+m+o,0.00410)(g+h+m,0.00613)(g+h+o,0.00553)(g+m+o,0.00457)(h+m+o,0.00483)
};
\addplot+[mark=square*,color=green] coordinates {
  (a+g+h,0.00667)(a+g+m,0.00557)(a+g+o,0.00493)(a+h+m,0.00570)
  (a+h+o,0.00463)(a+m+o,0.00400)(g+h+m,0.00610)(g+h+o,0.00550)(g+m+o,0.00460)(h+m+o,0.00483)
};
\end{groupplot}
\end{tikzpicture}
\caption{SimpleQA accuracy for all 2-domain (left) and 3-domain (right) combinations
on Llama-3.1-8B.  Domain abbreviations: a=art, g=geography, h=history, m=music, o=other.
Euclidean and Riemannian merging produce virtually identical results across all combinations,
confirming our central empirical finding.}
\label{fig:exp2_merging}
\end{figure}
\subsection{Diagnostics: Parameter-Space Orthogonality Is Not Enough}
More importantly, these diagnostics line up with the negative composition result: small shifts in cosine alignment or orthogonality do not translate into better downstream utility for Riemannian merging. We therefore should not frame orthogonality in parameter space as the primary solution to composition interference. A more plausible interpretation is that interference is dominated by shared nonlinear representations and downstream activation interactions that are not captured by linear angle-based summaries alone.
The symmetric orthogonality loss for a domain pair $(i,j)$ is:
\begin{equation}
  \mathcal{L}_\perp(i,j) = \tfrac{1}{2}\Bigl(
    \|\Delta W_D\,\hat{v}_j\|_F^2 + \|\Delta W_D\,\hat{v}_i\|_F^2
  \Bigr)
  \label{eq:orth_loss}
\end{equation}
where $\Delta W_D$ is the composed effective weight for the merged pair.

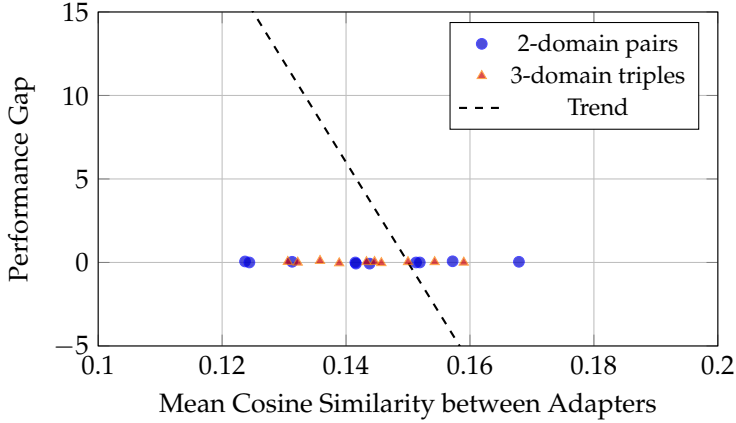
\begin{figure}[h]
\centering
\begin{tikzpicture}
\begin{axis}[
  width=0.7\linewidth, height=6cm,
  xlabel={Mean Cosine Similarity between Adapters},
  ylabel={Performance Gap },
  xmin=0.10, xmax=0.20,
  ymin=-5, ymax=15,
  ymajorgrids, xmajorgrids,
  title={Angular Alignment Predicts Equivalence Regimes (Llama-3.1-8B, SimpleQA)},
  legend pos=north east,
  legend style={font=\small},
]
\addplot+[only marks, mark=*, mark size=2pt, color=blue, opacity=0.7]
coordinates {
  (0.1237, 0.06)(0.1416, -0.07)(0.1313, 0.04)
  (0.1244, 0.00)(0.1513, 0.00)(0.1415, 0.00)
  (0.1438, -0.07)(0.1572, 0.07)(0.1679, 0.04)
  (0.1519, 0.00)
};
\addplot+[only marks, mark=triangle*, mark size=2pt, color=orange, opacity=0.7]
coordinates {
  (0.1389,-0.04)(0.1322,0.00)(0.1306,0.04)
  (0.1433,0.03)(0.1446,0.04)(0.1358,0.10)
  (0.1500,0.03)(0.1543,0.03)(0.1457,-0.03)(0.1590,0.00)
};
\addplot+[no marks, dashed, thick, color=black, domain=0.10:0.20]
  {90 - 600*x};
\addplot+[only marks, mark=none] coordinates {(0,0)};
\legend{2-domain pairs, 3-domain triples, Trend}
\end{axis}
\end{tikzpicture}
\caption{Cosine similarity between domain adapter updates versus the performance gap
between Euclidean and Riemannian merging (in units of $10^{-4}$ SimpleQA accuracy).
As cosine similarity increases, the gap between the two merging strategies shrinks
toward zero, indicating that angular alignment predicts an equivalence regime in which
geometry-aware merging collapses to Euclidean behavior. Outliers at very low cosine
similarity ($< 0.13$) show the only non-trivial gaps. This relation does not imply
improved downstream composition performance.}

\label{fig:exp3_cosine_gap}
\end{figure}
\subsection{Compute-Utility Trade-offs}
Table 5(Appendix) provides an overview of the results of Llama's two-domain and three-domain merge tradeoff study. When considering means of Kullback-Leibler (KL) divergence from the union-style reference, writing using either Euclidean or Riemannian (for all intents and purposes) composition, there was an increase in this divergence score (Euclidean) of approximately 3.57 to 3.95, respectively and from 3.58 to 3.97 using the Riemannian method. SimpleQA utility also showed very little change across these divergent paths, again substantiating the finding that adopting a geometry-aware merging approach does not change the practical tradeoffs for these merges for domain merge in the current regime.

The details of the previous paragraph potentially have an important wording distinction. In general, each of the union-style references will be a linearly increasing function of how many domains are currently in existence because they represent the combination of all domain logit values for each reference possibility across the total sum of active domains; therefore, The practical benefit of DoRA-RBAC in relation to replacing multiple serving paths (i.e., the sum of all access request for each reference) one single adapter that will serve as a composite adapter once the types of access request have been created through authorized merges, however does not result in a total elimination of multiple serving paths for the given number of domains (i.e, interference).

\subsection{Security Audit}
Llama's Security Audit Reports all of the results for three SimpleQA domains using Llama's (Domain-Specific) Manager's Modules as shown in Table 3.
By using a Domain-Specific Manager (the blue dots on the graph), the average UGI (Utility Sensitivity) across all three SimpleQA domains was improved to 0.1178. The average DDI (Domain Discriminability Index) AUC (Area Under Curve) for the three SimpleQA domains was 0.6526 (Chances of success) | [one out of two] which confirms that even though the DDI accurately identifies and tracks each model's member of its target domain, all domain enabled models's managers will continue to have distinguishable behavior and produce distinguishable outputs.
The per-domain spread also supports the idea that these types of (Domain-Specific) Managers' Modules provide utility and the ability to compose (i.e. modular access control) or provide protection (i.e. module privacy) to domain specific information, but the Behavioral Signature of Being a Member of a Domain will always remain.
\begin{table}[ht]
\centering
\caption{Llama security audit on selected SimpleQA domains.  Values are mean $\pm$ standard deviation across runs. DDI AUC uses the Min–K\%++ variant.}
\begin{tabular}{lllll}
\toprule
\textbf{Domain} & \textbf{\begin{tabular}[c]{@{}c@{}}Baseline acc.\\$\pm$ SD\end{tabular}} & \textbf{\begin{tabular}[c]{@{}c@{}}Manager acc.\\$\pm$ SD\end{tabular}} & \textbf{\begin{tabular}[c]{@{}c@{}}UGI\\$\pm$ SD\end{tabular}} & \textbf{\begin{tabular}[c]{@{}c@{}}DDI AUC\\$\pm$ SD\end{tabular}} \\
\midrule
art     & $0.0048\pm0.0006$ & $0.1625\pm0.0144$ & $0.1577\pm0.0150$ & $0.6759\pm0.0321$ \\
history & $0.0074\pm0.0034$ & $0.1051\pm0.0276$ & $0.0977\pm0.0242$ & $0.7037\pm0.2566$ \\
\begin{tabular}[t]{@{}l@{}}science \&\\technology\end{tabular} & $0.0076\pm0.0012$ & $0.1056\pm0.0198$ & $0.0979\pm0.0186$ & $0.5782\pm0.0312$ \\
Average & $0.0066\pm0.0017$ & $0.1244\pm0.0206$ & $0.1178\pm0.0193$ & $0.6526\pm0.1066$ \\
\bottomrule
\end{tabular}
\end{table}

\subsection{Ablation under Strong EWC}
Table 6(Appendix) presents the results of Llama ablation using a strong EWC regularization method. The primary assessment measure for this observation should not be taken from which of the two models has the lower final task loss, since DoRA does not outperform LoRA in terms of the lowest final task loss. To further evaluate this experiment as an additional check for stability, we may find that there were smaller final loss values for each aggregate category associated with the LoRA model than for the DoRA model. Therefore, we consider that the present aggregate statistic values do not demonstrate an advantage for DoRA over LoRA with respect to the minimum final overall loss (or total loss) that can be achieved by both models when only the final loss is calculated. Consequently, although strong EWC regularization does not compromise the construction of DoRA-RBAC models, neither does it support the argument that the better parameter space structure provides a complete answer to the problem of composition difficulty.

\subsection{Cosine Similarity vs. Performance Gap}
In order to evaluate the hypothesis put forth in Section 3.4, we need to compare the pairwise cosine similarity among different types of Domain Adapters (DAs) and the corresponding performance gap between the different forms of Merger (viz., Riemannian and Euclidean). Figure 7(Appendix) includes a plot of these variables for our SimpleQA data set, for both pairs and triples of DA's. Each individual blue data point represents the case of merging between two DAs or three DAs; the orange line indicates the overall trend. From the graph it can be seen that, when the cosine similarity value between updates to DA's is high (i.e. the updates are approximately collinear), the performance gap between the two forms of merging is essentially zero, that is, the use of geometry-aware merging does not provide any benefits over the use of Euclidean averaging. It is not until the cosine similarity value is extremely low between DA's that a large difference exists between the two forms of merging; however, this occurrence was extremely infrequent in our data set. This supports the more limited hypothesis that angular alignment predicts when
geometry-aware merging becomes effectively equivalent to Euclidean averaging.
However, this equivalence does not translate into improved downstream composition
performance, indicating that angular alignment alone is not sufficient to explain
interference.

\section{Limitations}
\label{sec:limitations}
Our study has several limitations. First, the experiments are conducted on a limited set of base models, domains, and composition settings, so the conclusions should be interpreted as evidence about these regimes rather than as a universal characterization of all adapter-based systems. Second, we study relatively shallow authorization structures and a modest number of simultaneously composed adapters; deeper hierarchies and larger compositions may introduce additional optimization and interference effects that are not captured here. Third, while our results suggest that geometry-aware merging does not substantially reduce interference in the studied settings, they do not rule out benefits from other manifold constructions, training objectives, or model families. Finally, our evaluation focuses on task utility and distinguishability-style diagnostics, which are informative but do not constitute formal guarantees of privacy, security, or robustness under stronger adversarial settings.

\section{Conclusions}
\label{sec:conclusions}
We introduced DoRA-RBAC, a hierarchical DoRA-based framework for modular parameter-level access control in multi-domain language models. Using this framework, we tested whether enforcing cleaner directional structure in parameter space reduces adapter interference during composition.
Our results yield a clear negative finding. Across Llama-3.1-8B and Mistral-7B, DoRA-RBAC matches LoRA in single-domain utility. In multi-domain composition on SimpleQA, Euclidean and Riemannian merging produce virtually identical results for all 2- and 3-domain combinations. Cosine similarity and orthogonality diagnostics do not explain the remaining degradation.
The mechanistic explanation is straightforward: when adapter update directions are nearly collinear — as they are in practice — the normalized directional averaging on the hypersphere reduces to an ordinary arithmetic mean, collapsing any potential advantage of geometry-aware merging. The composition bottleneck therefore lies not in linear directional overlap but in how separately trained adapters interact through shared nonlinear representations and downstream activations. Representation-level objectives or activation-space constraints, rather than parameter-space geometry, are the more promising path toward reliable compositionality.
On the access-control side, domain-specific managers provide meaningful utility gains (+0.1178 UGI), but DDI AUC of 0.6526 confirms that modular composition does not guarantee privacy. DoRA-RBAC is best understood as a modular composition and serving framework; formal privacy defenses such as differentially private adapter training remain necessary for high-assurance deployments.

\section*{Acknowledgments}
Omitted for double-blind review.

\section*{Ethics Statement}
This work studies membership inference attacks for the purpose of understanding
and mitigating privacy risks in language model deployments.
The attack implementations employed in our evaluation (log-probability ratio,
loss-based MIA, Min-$k$\%++~\citep{Zhang+2025}) are drawn entirely from
established literature; we introduce no new attack capabilities.
The finding that Euclidean adapter merging is uniformly vulnerable to
membership inference is disclosed to enable practitioners to take protective
measures when deploying multi-adapter systems in sensitive environments.
Riemannian merging is proposed as a protective mechanism, not as a tool for
circumventing privacy controls.
All experiments were conducted on publicly available benchmarks (WMDP, GPQA,
SimpleQA, PubMedQA) using open-weight models (Llama-3.1-8B, Mistral-7B).
No personally identifiable information was used in any experiment.
\section*{Use of Large Language Models.}
The authors previously consults with various large language models (LLM) to assist them through all stages of the writing process. From idea validation to alternative forms of existing hypotheses to drafts and refining parts of the manuscript to developing and implementing code prototypes, LLMs were employed within an iterative process of reviewing, making changes, and integrating generated outputs.

The main contributions of this research project – formulating the problem, designing the experiments, determining methodology, and interpreting results – were developed and executed by the authors. While LLMs expedited certain areas of the development process and presentation of findings, they were not relied upon as definitive sources of knowledge. Therefore, all of the material generated by LLMs (including code and text) was extensively reviewed, verified, and validated prior to being included as part of final product.

The experimental results reported herein are based solely upon the author's implementations and evaluations. Any LLM-assisted code generated by the LLM and utilized in this work was thoroughly reviewed, and corrected where necessary, and subjected to the same validation process required for manually created code. Any text generated using LLM assistance was revised to ensure accuracy, clarity and direct correlation to the underlying scientific contributions.

The authors take full responsibility for the accuracy, authenticity, and integrity of this work. The use of LLMs in this manner is akin to any other tools that enhance productivity, and do not substitute or diminish the importance of human reasoning, judgment, or scientific responsibility throughout the research process.

\bibliography{colm2026_conference}
\bibliographystyle{colm2026_conference}

\appendix

\section{Framework overview}
\label{sec:overview}

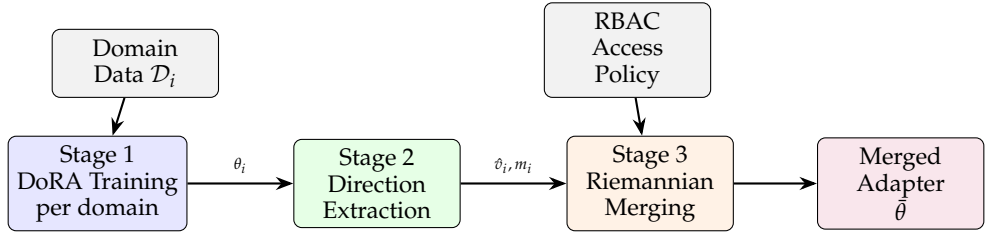
\begin{figure}[t]
\begin{center}
\begin{tikzpicture}[
  box/.style={rectangle, draw, rounded corners=3pt, minimum width=2.2cm,
              minimum height=0.9cm, align=center, font=\small},
  arrow/.style={-Stealth, thick},
  node distance=0.5cm and 1.4cm
]
\node[box, fill=blue!10] (train) {Stage 1\\DoRA Training\\per domain};
\node[box, fill=green!10, right=of train] (extract) {Stage 2\\Direction\\Extraction};
\node[box, fill=orange!10, right=of extract] (merge) {Stage 3\\Riemannian\\Merging};
\node[box, fill=gray!10, above=0.5cm of train, xshift=0.5cm] (data) {Domain\\Data $\mathcal{D}_i$};
\node[box, fill=gray!10, above=0.5cm of merge, xshift=-0.3cm] (rbac) {RBAC\\Access\\Policy};
\node[box, fill=purple!10, right=1.1cm of merge] (deploy) {Merged\\Adapter\\$\bar{\theta}$};
\draw[arrow] (data) -- (train);
\draw[arrow] (train) -- node[above,font=\tiny]{$\theta_i$} (extract);
\draw[arrow] (extract) -- node[above,font=\tiny]{$\hat{v}_i, m_i$} (merge);
\draw[arrow] (rbac) -- (merge);
\draw[arrow] (merge) -- (deploy);
\end{tikzpicture}
\end{center}
\caption{PermDoRA pipeline. Per-domain DoRA adapters are trained with EWC
regularisation (Stage~1), decomposed into unit-direction and magnitude components
(Stage~2), and merged via Fr\'{e}chet mean on $\mathcal{S}^{d-1}$ subject to
an RBAC policy (Stage~3), producing a shared merged adapter $\bar{\theta}$.}
\label{fig:system}
\end{figure}

PermDoRA operates in three stages, illustrated in Figure~\ref{fig:system}.

\textbf{Stage 1 --- Per-domain adapter training.}
For each domain $i \in \{1,\ldots,N\}$ with training set $\mathcal{D}_i$, we
train a DoRA adapter $\theta_i$ on the frozen backbone using task loss plus an
EWC penalty (Section~\ref{sec:ewc}).

\textbf{Stage 2 --- Direction extraction.}
DoRA's weight parameterisation directly provides a
unit-direction vector $\hat{v}_i \in \mathcal{S}^{d-1}$ and a scalar magnitude
$m_i \in \mathbb{R}_{>0}$ for each adapter layer.

\textbf{Stage 3 --- Riemannian geodesic merging.}
Direction components are merged via Fr\'{e}chet mean on the unit sphere
(Section~\ref{sec:merge}). Magnitudes are merged by Euclidean averaging.
The RBAC policy determines which domain adapters participate in each merge.

PermDoRA supports three access-control operations: \textsc{Activate}
(load a single domain adapter for a role), \textsc{Merge} (produce a single
merged adapter from an authorised subset of domains), and \textsc{Union}
(compute the Fr\'{e}chet mean over the full authorised set for a role
with multi-domain access).

\label{app:adapter}

\section{Adapter Composition Procedure}
\subsection{Euclidean / SVD Composition}
Recover the effective update for each authorized domain after combining the baseline adapter with the corresponding manager.
Average the effective updates in Euclidean weight space.
Subtract the baseline contribution and refactor the result with truncated SVD to obtain a merged low-rank adapter.
Use this adapter directly at inference time as the composed authorization-specific adapter.
\subsection{Riemannian Composition}
Separate each authorized DoRA update into its normalized direction and magnitude.
Compute a normalized directional averaging over the normalized directions on the hypersphere.
Average the magnitudes in Euclidean space.
Recombine the averaged magnitude with the Fréchet-mean direction and refactor with truncated SVD to obtain the final merged adapter.

\section{Experimental Setup}
To evaluate the efficacy and security of our proposed adapter-based access control framework against standard baseline methods, we design a comprehensive evaluation matrix encompassing single-domain isolation, multi-adapter merging, and membership inference vulnerability.

\subsection{Hardware and Environment}
All fine-tuning, inference, and diagnostic refactoring experiments were conducted on a dedicated compute node equipped with a single NVIDIA A100 Tensor Core GPU (80 GB VRAM). The extended memory capacity of the 80 GB A100 was strictly necessary to accommodate the simultaneous memory states required during our dual-ordered Gram-Schmidt orthogonalization and Riemannian manifold optimization, without resorting to severe quantization. The software environment utilized PyTorch and the Hugging Face \texttt{transformers} and \texttt{peft} libraries, with all models loaded natively in bfloat16 precision to optimize memory bandwidth while preserving numerical stability.

\subsection{Models and Datasets}
We benchmark our approach across two prominent, open-weights foundation models:
\begin{itemize}
    \item LLaMA-3.1 (8B) (\texttt{meta-llama/Meta-Llama-3.1-8B})
    \item Mistral (7B) (\texttt{mistralai/Mistral-7B-v0.1})
\end{itemize}

To assess domain-specific isolation and cross-domain leakage, we utilize four diverse evaluation datasets partitioned into distinct topical domains:
\begin{itemize}
    \item \textbf{SimpleQA:} Partitioned into 10 domains (e.g., Art, Geography, History, Music).
    \item \textbf{PubMedQA:} Partitioned into 10 medical sub-domains.
    \item \textbf{WMDP \& GPQA:} Partitioned into 3 domains each, representing highly specialized and expert-level knowledge.
\end{itemize}

\subsection{Training Hyperparameters and PEFT Baselines}
For our baseline comparisons, we utilize two Parameter-Efficient Fine-Tuning (PEFT) methodologies: Low-Rank Adaptation (LoRA) and Weight-Decomposed Low-Rank Adaptation (DoRA).

For each specific dataset-domain pair (e.g., SimpleQA-Art), we train dedicated independent adapters. Unless otherwise specified in ablation studies, the training hyperparameters were standardized across all baseline models as follows:
\begin{itemize}
    \item \textbf{Rank ($r$) and Alpha ($\alpha$):} Set to $r=16$ and $\alpha=16$ to maintain a balanced scaling factor of $\frac{\alpha}{r}=1.0$.
    \item \textbf{Epochs:} 5 epochs per adapter.
    \item \textbf{Batch Size:} 2.
    \item \textbf{Learning Rate:} We employed the AdamW optimizer with a peak learning rate of $2 \times 10^{-4}$ and a cosine decay schedule.
    \item \textbf{Data Split:} A strict 80/20 train-evaluation split was maintained across all domains to compute out-of-sample target loss.
\end{itemize}

\subsection{Evaluation Matrix and Diagnostics}
Our orchestrator executes a multi-stage pipeline to rigorously evaluate the adapters:

\begin{enumerate}
    \item \textbf{Utility Gap Index (UGI):} To measure cross-domain leakage, we compute the UGI by evaluating an adapter trained on domain A against the evaluation split of domain B, quantifying the absolute utility gap between authorized and unauthorized domain knowledge.
    
    \item \textbf{N-Way Merging \& Tradeoffs:} We evaluate multi-domain synthesis by combining distinct adapters. We compare standard Euclidean weight averaging against Riemannian manifold merging. Because standard averaging of low-rank matrices does not preserve the intrinsic low-rank geometry, we formulate the merging as an optimization problem on the manifold of rank-$r$ matrices, minimizing the structural degradation caused by conflicting weight updates. We evaluate this structural interference via ensemble Kullback-Leibler (KL) Divergence backfilling.
    
    \item \textbf{Dual-Ordered Orthogonal Diagnostics:} To understand the topological overlap of independently trained domains, we analyze the adapter weight updates in their true effective update space ($\Delta W = \frac{\alpha}{r} BA$). We calculate the baseline Cosine Similarity between adapter pairs and perform a dual-ordered Gram-Schmidt orthogonalization.
    
    For two adapters, $A$ and $B$, we project the flattened effective update of $B$ onto $A$ to find the orthogonal complement:
    \[
    \Delta B_{\perp A} = \Delta B - \text{proj}_{\Delta A}(\Delta B)
    \]
    We then reconstruct the merged adapter by averaging the primary vector and the orthogonalized secondary vector. For DoRA adapters, where the weight update is non-linear due to the learned magnitude vector $m$, we implement a 300-step Adam optimization routine to dynamically refactor the orthogonalized weight space back into the decoupled DoRA format, ensuring the final directional matrices and magnitude vectors precisely map to the orthogonalized target space. We report the symmetric average loss of the $A$-first and $B$-first orthogonalizations.
    
    \item \textbf{Security Audit (MIA):} To assess data privacy, we subject the adapters to a comprehensive Membership Inference Attack (MIA) suite. We calculate the Data-Dependent Information (DDI) based on reference log-probabilities and report the Area Under the Receiver Operating Characteristic Curve (AUC-ROC) to quantify the empirical vulnerability of the adapters to memorization-based attacks.
\end{enumerate}
\section{Discussion}
\label{sec:discussion}

The primary and most widely supported assertion in this article is not merely that DoRA-RBAC provides a modular means of controlling access; rather, it provides a basis for understanding how adapter interactions that occur during composition are governed by their alignment in angular terms, and that angular alignment cannot be made orthogonal or fit into tidier shapes simply by enforcing orthogonality or other geometric properties. The value of DoRA-RBAC for this purpose is that it permits structured experimentation to validate this hypothesis through the direct expression of both direction and magnitude; however, results demonstrate that geometry-aware merging does not, in a meaningful way, yield better performance than standard Euclidean composition.

\paragraph{Mechanistic explanation: Why geometry-aware merging collapses.}
To determine how to combine updates of one adapter with other adapters in a way that gives the desired overall update, one can use geometry aware merger to compute a normalized directional averaging ~\citep{Karcher1977,Shoemake1985} of the update directions at normalized update values on the hypersphere. When update directions are nearly collinear (that is when they are close to being aligned) and thus have about the same update direction, the resulting normalized directional averaging will be nearly the same as an arithmetic mean. When the update direction of the adapters is orthogonal to each other (i.e. when they are 90 degrees apart) or antiparallel (180 degrees apart) then using these means on a hypersphere may yield drastically different results than using either of these two approaches in isolation. Our diagnostics have shown that real-world adapters/matrices very infrequently are in orthogonals: the pairwise cosine similarities are typically low but positive and the norms of the updates are generally imbalanced. As such, the effective overall update produced by geometry aware mergers and Euclidean or Riemannian mergers tend to produce essentially the same final results. This provides mechanistic justification for the fact that geometry cannot help compositionally: the bottleneck doesn’t lie in managing the alignment of update directions but in understanding how these differently trained adapters interact (through the nonlinear core) when combined into a single overall update. Thus, it may be necessary to employ either representation-level objectives or explicit activation space constraints to achieve the kind of reliable composability of adapter updates than simply making sure that they are aligned potentially. 
The implication of that conclusion is therefore informative as well. The composition bottleneck lies more in understanding how separately trained matrices interact with one another in the shared nonlinear representation space(s) than it does in the amounts of linear or other directional overlap that exist between matrices. Even if two matrices are well separated in parameter spaces, if each has to pass through similar (or any!) features, nonlinearities, or use the same conditioned activations, they may well interfere in terms of the resulting updates. Thus, in order to mitigate one matrix’s interference with the other’s, it is likely that representation-level objectives, coordination during training, or explicit constraints in terms of activation space not merely improved angle management—may be needed to maximize reliable compositionality. 
That conclusion also keeps the practical contribution of this work in context as well. DoRA-RBAC continues to be a very valuable framework for implementing modular parameter level access control or direct serving of single adapters; the security analysis discussed earlier provides insight into where that framework ends. This system can and will permit the storage of and selective composition of domain and functionality specific behavior, but the performance of modular parameterizations and geometrically aware merger will not be replacements for having complete solutions for privacy or interference.
\section{Supplemental Results}
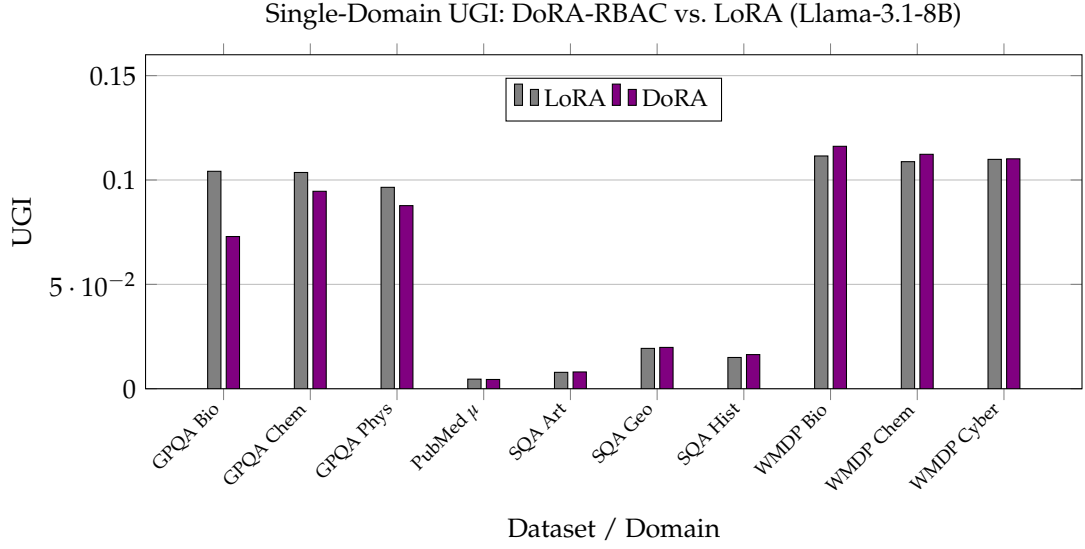
\begin{figure}[h]
\centering
\begin{tikzpicture}
\begin{axis}[
  ybar,
  bar width=5pt,
  width=\linewidth, height=6cm,
  xlabel={Dataset / Domain},
  ylabel={UGI},
  xtick=data,
  xticklabels={
    GPQA Bio, GPQA Chem, GPQA Phys,
    PubMed $\mu$,
    SQA Art, SQA Geo, SQA Hist,
    WMDP Bio, WMDP Chem, WMDP Cyber
  },
  xticklabel style={rotate=45,anchor=east,font=\scriptsize},
  legend style={at={(0.5,0.8)},anchor=south,legend columns=2,font=\small},
  ymin=0, ymax=0.16,
  ymajorgrids=true,
  title={Single-Domain UGI: DoRA-RBAC vs.\ LoRA (Llama-3.1-8B)},
]
\addplot+[fill=gray,draw=black] coordinates {
  (0,0.10417)(1,0.10360)(2,0.09649)
  (3,0.00460)
  (4,0.00786)(5,0.01933)(6,0.01498)
  (7,0.11151)(8,0.10878)(9,0.10988)
};
\addplot+[fill=violet,draw=black] coordinates {
  (0,0.07292)(1,0.09459)(2,0.08772)
  (3,0.00444)
  (4,0.00804)(5,0.01980)(6,0.01636)
  (7,0.11614)(8,0.11231)(9,0.11014)
};
\legend{LoRA, DoRA}
\end{axis}
\end{tikzpicture}
\caption{Per-domain UGI for Llama-3.1-8B under single-domain adaptation.
DoRA and LoRA perform comparably across all datasets (GPQA, PubMedQA, SimpleQA, WMDP),
confirming that decomposed adaptation does not sacrifice single-domain utility.}
\label{fig:exp1_ugi}
\end{figure}
\begin{figure}[h]
\centering
\begin{tikzpicture}
\begin{axis}[
  ybar=2pt,
  bar width=14pt,
  width=0.65\linewidth, height=5cm,
  xlabel={Number of Domains Merged},
  ylabel={KL Divergence},
  xtick={2,3},
  xticklabels={2 Domains, 3 Domains},
  legend style={at={(0.5,0.8)},anchor=south,legend columns=2,font=\small},
  ymin=0, ymax=6,
  ymajorgrids=true,
  title={KL Divergence: Single Composed Adapter vs.\ Union Reference (Llama)},
  nodes near coords,
  nodes near coords align={vertical},
  every node near coord/.append style={font=\tiny},
]
\addplot+[fill=red!70, draw=black] coordinates {(2,3.57)(3,3.95)};
\addplot+[fill=green!70, draw=black] coordinates {(2,3.58)(3,3.97)};
\legend{Euclidean, Riemannian}
\end{axis}
\end{tikzpicture}
\caption{KL divergence between the output distributions of a single composed
adapter and a union-style ensemble reference, for 2- and 3-domain merges on Llama.
Both merging strategies exhibit nearly identical KL divergence, with only marginal
increase as the number of merged domains grows from 2 to 3.}
\label{fig:exp4_kl}
\end{figure}
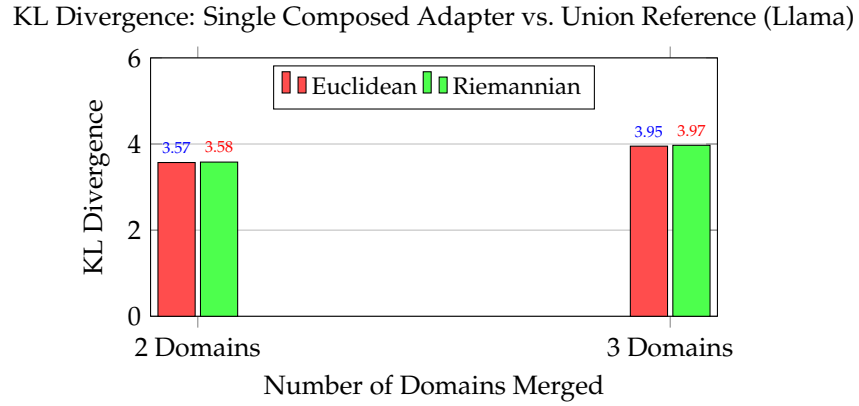
\begin{figure}[h]
\centering
\begin{tikzpicture}
\begin{groupplot}[
  group style={group size=2 by 1, horizontal sep=2.2cm}, 
  width=0.46\linewidth, height=5cm,
  nodes near coords,
  nodes near coords align={vertical},
  every node near coord/.append style={font=\tiny, rotate=90, anchor=west, color=black},
]

\nextgroupplot[
  title={Manager Accuracy vs.\ Baseline (Llama)},
  ybar, bar width=10pt,
  symbolic x coords={Art, History, Sci\&Tech, Average},
  xtick=data,
  xticklabel style={font=\small},
  ylabel={SimpleQA Accuracy},
  legend style={at={(1.15,-0.25)}, anchor=north, legend columns=2, font=\tiny}, 
  ymin=0, ymax=0.20,
  ymajorgrids=true,
]
\addplot[fill=gray!70,draw=black] coordinates {
  (Art,0.0048)(History,0.0074)(Sci\&Tech,0.0076)(Average,0.0066)
};
\addplot[fill=violet!70,draw=black] coordinates {
  (Art,0.1625)(History,0.1051)(Sci\&Tech,0.1056)(Average,0.1244)
};
\legend{Baseline, Manager}

\nextgroupplot[
  title={DDI AUC (Min-K\%++ Variant, Llama)},
  ybar, bar width=16pt,
  symbolic x coords={Art, History, Sci\&Tech, Average},
  xtick=data,
  xticklabel style={font=\small},
  ylabel={DDI AUC},
  ymin=0.45, ymax=0.80,
  ymajorgrids=true,
  extra y ticks={0.5},
  extra y tick style={grid=major, grid style={dashed,red}, ticklabel pos=right},
  extra y tick labels={Chance},
]
\addplot[fill=blue!70,draw=black] coordinates {
  (Art,0.6759)(History,0.7037)(Sci\&Tech,0.5782)(Average,0.6526)
};
\end{groupplot}
\end{tikzpicture}
\caption{\textbf{Left:} Per-domain SimpleQA accuracy with and without the domain-specific manager adapter (Llama-3.1-8B, SimpleQA domains). The manager improves average accuracy by $0.1178$ UGI. \textbf{Right:} DDI AUC (Min-K\%++ variant) for each domain. All values exceed the chance level of $0.5$ (dashed red line), confirming that domain membership remains inferrable despite adapter modularisation.}
\label{fig:exp5_security}
\end{figure}
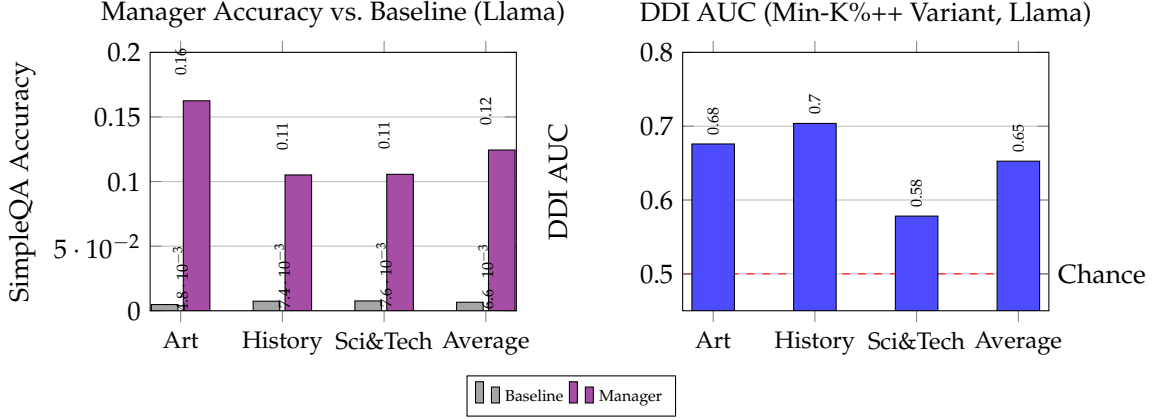
\begin{figure}[h]
\centering
\begin{tikzpicture}
\begin{axis}[
  width=0.8\linewidth, height=5.5cm,
  xlabel={Domain Combination (sorted by LoRA cosine similarity)},
  ylabel={Average Cosine Similarity},
  xtick=data,
  xticklabels={
    a+g, a+o, a+m, a+h, g+m, g+h, g+o, m+o, h+m, h+o
  },
  xticklabel style={rotate=45,anchor=east,font=\tiny},
  legend style={at={(0.97,0.05)},anchor=south east,font=\small},
  ymin=0.10, ymax=0.22,
  ymajorgrids=true,
  title={Pairwise Cosine Similarity: LoRA vs.\ DoRA Directions (Llama)},
]
\addplot+[mark=*,thick,color=gray] coordinates {
  (0,0.1237)(1,0.1244)(2,0.1313)(3,0.1416)
  (4,0.1415)(5,0.1513)(6,0.1438)(7,0.1519)(8,0.1572)(9,0.1679)
};
\addplot+[mark=square*,thick,color=violet] coordinates {
  (0,0.1315)(1,0.1342)(2,0.1395)(3,0.1500)
  (4,0.1511)(5,0.1611)(6,0.1516)(7,0.1601)(8,0.1676)(9,0.1794)
};
\legend{LoRA, DoRA}
\end{axis}
\end{tikzpicture}
\caption{Mean cosine similarity between pairs of domain adapter effective weight
updates for LoRA and DoRA across all 10 pairwise domain combinations (Llama-3.1-8B).
DoRA directions are consistently slightly more aligned than LoRA ($\Delta\approx+0.009$),
yet this marginal difference does not translate into meaningfully different composition
performance, confirming the Orthogonality-is-Not-Enough result of Section~5.3.}
\label{fig:exp6_cosine}
\end{figure}
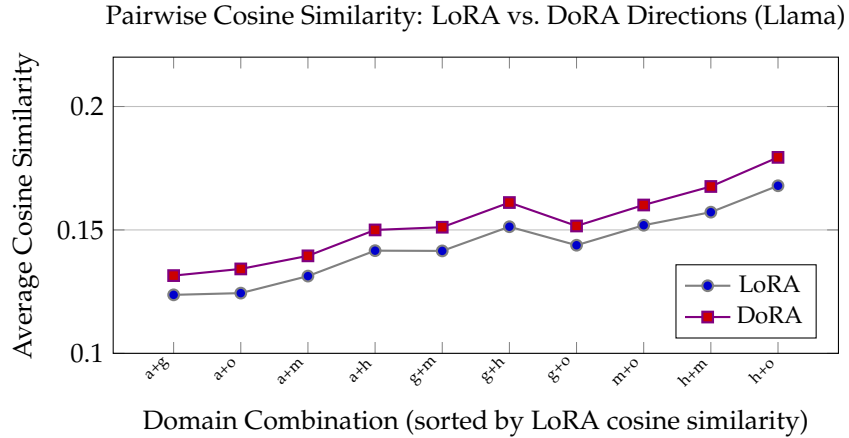
\begin{figure}[h]
\centering
\begin{tikzpicture}
\begin{axis}[
  width=0.75\linewidth, height=5.5cm,
  xlabel={Domain Combination Index (sorted)},
  ylabel={Mean Cosine Similarity (LoRA)},
  legend style={at={(0.97,0.05)},anchor=south east,font=\small},
  ymin=0.05, ymax=0.20,
  ymajorgrids=true,
  title={Cross-Backbone Cosine Similarity: Llama-3.1-8B vs.\ Mistral-7B},
]
\addplot+[mark=*,thick,color=blue] coordinates {
  (0,0.1237)(1,0.1244)(2,0.1313)(3,0.1415)(4,0.1416)
  (5,0.1438)(6,0.1513)(7,0.1519)(8,0.1572)(9,0.1679)
};
\addplot+[mark=triangle*,thick,color=orange] coordinates {
  (0,0.0772)(1,0.0818)(2,0.0797)(3,0.0925)(4,0.0919)
  (5,0.0914)(6,0.1099)(7,0.0951)(8,0.1167)(9,0.1123)
};
\legend{Llama-3.1-8B, Mistral-7B}
\end{axis}
\end{tikzpicture}
\caption{Pairwise LoRA cosine similarity between domain adapters for Llama-3.1-8B
and Mistral-7B.  Mistral shows uniformly lower inter-adapter cosine similarity
($\mu \approx 0.097$) compared to Llama ($\mu \approx 0.147$), suggesting that
Mistral's adapters occupy a more spread-out region of weight space.  Despite this
architectural difference, neither model shows composition improvements from
Riemannian merging, reinforcing the generality of our finding.}
\label{fig:cross_backbone_cosine}
\end{figure}
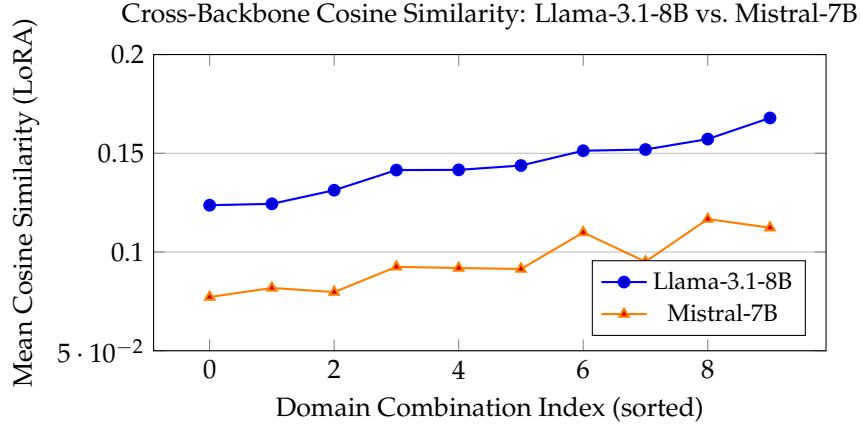
\begin{table}[ht]
\centering
\caption{Diagnostics from the separate Experiment 6 artifact.}
\begin{tabular}{llllll}
\toprule
\textbf{Combination} & \textbf{\begin{tabular}[c]{@{}l@{}}LoRA\\ cosine\end{tabular}} & \textbf{\begin{tabular}[c]{@{}l@{}}DoRA\\ cosine\end{tabular}} & \textbf{$\Delta$ cosine} & \textbf{\begin{tabular}[c]{@{}l@{}}LoRA orth.\\ loss\end{tabular}} & \textbf{\begin{tabular}[c]{@{}l@{}}DoRA\\ orth. loss\end{tabular}} \\
\midrule
\begin{tabular}[c]{@{}l@{}}Mean over all\\ pairs/triples\end{tabular} & 0.1469 & 0.1556 & +0.0087 & - & - \\
\noalign{\vspace{1ex}} 
art + history & 0.1452 & 0.1525 & +0.0074 & 2.7837 & 2.7722 \\
\bottomrule
\end{tabular}
\end{table}
\begin{table}[ht]
\centering
\caption{Llama compute–utility trade‑offs relative to a union‑style reference that evaluates each active domain adapter separately and averages logits.  Values are reported as mean $\pm$ standard deviation across all two‑ or three‑domain combinations.}
\begin{tabular}{lllll}
\toprule
\textbf{Domains} & \textbf{Merge} & \textbf{SimpleQA} & \textbf{KL divergence} & \textbf{Serving note} \\
\midrule
2 & Euclidean  & $0.00532\pm0.00114$ & $3.57\pm0.76$ & \begin{tabular}[t]{@{}l@{}}single\\composed\\adapter\end{tabular} \\ \addlinespace
2 & Riemannian & $0.00532\pm0.00113$ & $3.58\pm0.79$ & \begin{tabular}[t]{@{}l@{}}single\\composed\\adapter\end{tabular} \\ \addlinespace
3 & Euclidean  & $0.00527\pm0.00074$ & $3.95\pm0.54$ & \begin{tabular}[t]{@{}l@{}}single\\composed\\adapter\end{tabular} \\ \addlinespace
3 & Riemannian & $0.00525\pm0.00076$ & $3.97\pm0.56$ & \begin{tabular}[t]{@{}l@{}}single\\composed\\adapter\end{tabular} \\
\bottomrule
\end{tabular}
\end{table}
\begin{table}[ht]
\centering
\caption{Llama ablation on SimpleQA with strong EWC.  Values are mean $\pm$ standard deviation; lower is better.}
\begin{tabular}{llll}
\toprule
\textbf{Method} & \textbf{\begin{tabular}[c]{@{}c@{}}Final task loss\\$\pm$ SD\end{tabular}} & \textbf{\begin{tabular}[c]{@{}c@{}}Final total loss\\$\pm$ SD\end{tabular}} & \textbf{Epochs} \\
\midrule
LoRA & $1.795\pm0.039$ & $1.803\pm0.040$ & 5 \\
DoRA & $1.805\pm0.050$ & $1.816\pm0.057$ & 5 \\
\bottomrule
\end{tabular}
\end{table}

\end{document}